\documentclass[a4paper,fleqn]{cas-sc}

\usepackage[numbers]{natbib}
\usepackage{siunitx}
\usepackage[hang,small,bf,tight]{subfigure}
\usepackage[justification=centering]{caption}
\usepackage{subcaption}
\usepackage{graphicx}
\usepackage{multirow}
\usepackage{amsmath}
\usepackage{graphics}
\usepackage{float}
\restylefloat{table}

\begin{document}
\let\WriteBookmarks\relax
\def\floatpagepagefraction{1}
\def\textpagefraction{.001}

\shorttitle{Finding RoIs in WSI using MIL}

\shortauthors{M. Afonso et~al.}

\title [mode = title]{Finding Regions of Interest in Whole Slide Images Using Multiple Instance Learning}                      


%
\author[1]{Martim Afonso}[]
\ead{martim.afonso@tecnico.ulisboa.pt}





\affiliation[1]{organization={Instituto Superior Técnico, Universidade de Lisboa},
    addressline={Av. Rovisco Pais}, 
    city={Lisbon},
    postcode={1049-001}, 
    country={Portugal}}

\affiliation[2]{organization={Division of Cancer Epidemiology and Genetics, National Cancer Institute, National Institutes of Health},
    city={Bethesda},
    postcode={20850}, 
    state={Maryland},
    country={USA}}
    
\affiliation[3]{organization={INESC-ID},
    addressline={R. Alves Redol 9}, 
    city={Lisbon},
    postcode={1000-029}, 
    country={Portugal}}


\author[2]{Praphulla M.S. Bhawsar}[]
\author[2]{Monjoy Saha}[]

\author[2]{Jonas S. Almeida}[]

\author[1,3]{Arlindo L. Oliveira}[]





\begin{abstract}
Whole Slide Images (WSI), obtained by high-resolution digital scanning of microscope slides at multiple scales, are the cornerstone of modern Digital Pathology. However, they represent a particular challenge to AI-based/AI-mediated analysis because pathology labeling is typically done at slide-level, instead of tile-level. It is not just that medical diagnostics is recorded at the specimen level, the detection of oncogene mutation is also experimentally obtained, and recorded by initiatives like The Cancer Genome Atlas (TCGA), at the slide level. This configures a dual challenge: a) accurately predicting the overall cancer phenotype and b) finding out what cellular morphologies are associated with it at the tile level. To address these challenges, a weakly supervised Multiple Instance Learning (MIL) approach was explored for two prevalent cancer types, Invasive Breast Carcinoma (TCGA-BRCA) and Lung Squamous Cell Carcinoma (TCGA-LUSC). This approach was explored for tumor detection at low magnification levels and TP53 mutations at various levels. Our results show that a novel additive implementation of MIL matched the performance of reference implementation (AUC 0.96), and was only slightly outperformed by Attention MIL (AUC 0.97). More interestingly from the perspective of the molecular pathologist, these different AI architectures identify distinct sensitivities to morphological features (through the detection of Regions of Interest, RoI) at different amplification levels. Tellingly, TP53 mutation was most sensitive to features at the higher applications where cellular morphology is resolved.
\end{abstract}




\maketitle
\section{Introduction}

Whole slide imaging is the automated process of digitally scanning whole microscope slides with high resolution. During this process, images from each field of view at different resolutions are taken and joined together to create a single digital image pyramid file, 
known as a whole slide image (WSI)~\cite{hanna2020whole}. 
This digital format, supported by a number of standard serializations, facilitates their distribution for diagnostic, education, and research purposes~\cite{Dimitriou2019}. WSIs
play a critical role in cancer diagnosis~\cite{Hou2022}.

Deep learning has been particularly successful in medical imaging applications such as diagnosis~\cite{Ilse2018}, 
sub-type classification~\cite{Lu2021, Hou2022} and prognosis~\cite{Zhu2017}.
However, deep learning of WSI faces three main challenges~\cite{khened2021generalized}: handling large image dimensionality at multiple scales; the lack of strongly annotated data; and more generally the difficulties inherent to approaching classification with information retrieval. 

The first factor, the large dimension of the images across multiple scales in the WSI image pyramid starts with resolutions at the base in the order of 100,000 $\times$ 100,000 pixels. The sheer size makes it difficult to feed the images directly as input to computer vision models. To overcome this issue, slides are usually divided into multiple fixed-size patches (also described as tiles) that are then used as input to the models.

The second factor is that the labeling is performed at slide level (for the whole WSI, not for individual tiles) or even at the patient level (one label per patient,  which can have multiple slides). Expert labeling at the pixel level would be costly and exhausting to the Pathologist. Wihtout pixel-level annotations, fully-supervised approaches cannot be directly employed and require instead weakly-supervised~\cite{Wang2020, Ilse2018, Li2020} or unsupervised~\cite{nayak2013classification, sheikh2022unsupervised} approaches such as multiple instance learning, MIL \cite{dietterich}, used in the work reported here.

Finally, it can be challenging to retrieve the relevant pathology information from relatively unstructured clinical reports. As a consequence, the interpretability of regions of interest (RoI) identified by deep learning models is often poor to the point that attempting explanation at the morphology level can only be approached as an exploratory exercise. However, the Regions of interest approach, where recurring morphological patterns are observed, is a familiar procedure in Digital Pathology, where it is sometimes described as “virtual staining”. Accordingly, in the work described here, we “stain” the raw images with heatmaps produced using the deep-learned activation scores.

\section{Materials and Methods}

\subsection{Multiple Instance Learning}

Multiple instance learning (MIL)~\cite{dietterich} is a weakly supervised learning approach, where instances are grouped into sets called bags. A label is assigned to the entire bag, while the individual instance labels remain unknown. The standard MIL definition uses a bag of instances as $X=\{x_1,...,x_K\}$ that do not have dependency nor ordering among each other.  We will also assume that the size of the bag K might vary for different bags. 
For each bag $X$, there is a binary label $Y$ $\in$ \{0,1\}. Each instance within the bag has a label $y_i$, where $y_i$ $\in$ \{0,1\}. 
However, the individual instance labels are unknown and are not accessible during training. 
Thus, the MIL problem can be written as the following:
\begin{equation} \label{eq1}
    Y = 
    \begin{cases}
      0, \quad \Sigma_{k} y_k = 0 \\
      1, \quad Otherwise 
    \end{cases}\,.
\end{equation}
In other words, we assume that all negative bags contain only negative instances and that positive bags contain at least one positive instance. To classify a bag as positive we only have to consider one instance as positive.

This standard assumption also posits that models must be permutation-invariant (there is no order nor dependency among instances). More specifically, 
it needs an aggregation operator that is permutation-invariant, usually referred to as a MIL pooling operator.

In a more general sense, a MIL model for bag classification can be expressed by a 3-step process~\cite{Ilse2018}:
\begin{itemize}
\item A transformation of the instances with a function $f$;
\item A permutation-invariant function $\sigma$ that combines all instance transformations into a final bag representation;
\item A final transformation $g$ that receives the bag representation and outputs a final bag score.
\end{itemize}

In the case of WSI analysis, we can consider each slide as a bag that contains several patches as instances. We have slide-level labels (bag label), but we do not have pixel/patch-level labels (instance labels).

To provide some degree of interpretability, as well as better results, a variation of the attention mechanism can be used as a MIL pooling operator. Because it acts as a weighted average of instances, the original definition can be adapted to be permutation-invariant, making it a valid MIL pooling operator. These 
attention scores can be used to build a heatmap that allows for the interpretation of which parts of an image are responsible for the final classification. There have been multiple works that use variations of the attention mechanism with the MIL framework in order to achieve this~\cite{guo2023robust, Hou2022, Konstantinov2022, Li2021, Li2020, Lu2021, lu2021data, Rymarczyk2021, Yao2020, Zhao2022}. 
\subsection{Model Architectures}
Due to the need for interpretability of the results, when choosing and building the models,  we focused on MIL models that use an attention pooling mechanism. Since we also prioritized efficiency, we chose models with a relatively simple architecture. Specifically, we focused on the original attention MIL (AMIL), proposed by Ilse et al.~\cite{Ilse2018} and the additive MIL (AdMIL), proposed by Javed et al. ~\cite{javed2022additive}.

Ilse et al. proposed a weighted average of the instances as a MIL pooling operator, where the weights were trained on a two-layered neural network. This operator corresponds  to a version of the attention mechanism where all instances are independent.

In this approach, we have a 3-part model $g$, defined as
\begin{equation} 
   g(x) = (p \; \circ \; a \; \circ \; f)(x),
\end{equation}
where $f$ is the feature extractor for each instance $x_i$ of the bag $x$ with $n$ instances, that returns the respective embedding $h_i$, where $h_i \in h = [h_1, ... , h_n]$:
\begin{equation}
    h_i = f(x_i).
\end{equation}

$a$ is the attention module, defined as:
\begin{equation}
    a_i(h_i) = softmax_i(\psi_a(h))h_i
\end{equation}
and $p$ is the predictor function or bag classifier:
\begin{equation}\label{eq10}
    p(h) = \psi_p(\sum_{i=1}^{n} a_i(h_i)).
\end{equation}
Both $\psi_a$ and $\psi_p$ are multi-layer perceptrons.

The attention mechanism is similar to Badhanau's attention \cite{Bahdanau2015}, with the main difference that the instances 
are considered sequentially independent:
\begin{equation}
    \psi_a(h_i) = w^Ttanh(Vh_i^T)
\end{equation}
with both $w$ and $V$ as trainable parameters.

However, this approach still presents some problems, regarding what is being 
interpreted in the image, because the attention scores might not be expressive enough 
for the explainability of the results. Javed et al.~\cite{javed2022additive} noted that the non-linear relationship between the attention values 
and the final prediction makes the visual interpretation inexact and incomplete.

The argument is that the attention scores obtained are not necessarily representative of tumors, 
but might still be needed for the downstream prediction, meaning they might be necessary but not 
sufficient for the final classification. The scores might indicate that the patch provides strong positive or 
negative evidence for the final prediction, but it does not make a distinction between 
the two. Moreover, the attention scores do not consider the contribution of multiple patches. Two patches 
might have moderate scores, but together they might provide strong evidence for the final 
prediction.

To address these problems, Ilse et al. proposed a new additive method for better visual 
interpretability and patch classification. To better specify the contributions 
from each patch, the final prediction function of the model described in equation \ref{eq10} 
undergoes the following change:

\begin{equation}
   p(h) = (\sum_{i=1}^{n} \psi_p(a_i(h_i))),
\end{equation}
where $\psi_p$ produces a score for each patch. With this change, $\psi_p(a_i(x))$ becomes the 
contribution for patch $i$ in a bag, turning it into a more accurate proxy of an instance classifier.
The patch scores are then added to produce a final bag-level score, which in turn is 
converted to a final probability distribution using a $softmax$, to obtain the final
classification.

For interpretability and slide inpainting, the patch scores obtained can be passed through a $sigmoid$ function to produce a bounded patch contribution score between 0 and 1, where 0 - 0.5 values 
represent inhibitory scores and 0.5 - 1 values represent excitatory scores.

AMIL was originally trained with each WSI patch as bag (where each instance is a portion of that patch) for tumor detection, while AdMIL was trained for cancer sub-type classification and metastasis detection. The application of these models for tasks such as tumor detection and gene mutation detection can give provide insights into the models' ability to generalize for different medical tasks. Furthermore, the differences in their employment of the MIL framework make it worthwhile to study and understand how these models focus on the different regions of slides.

\begin{figure}[htbp]
    \centering
           \subfigure[Attention MIL Architecture.]{\label{fig:AttentionMIL}\includegraphics[width=0.25\textwidth]{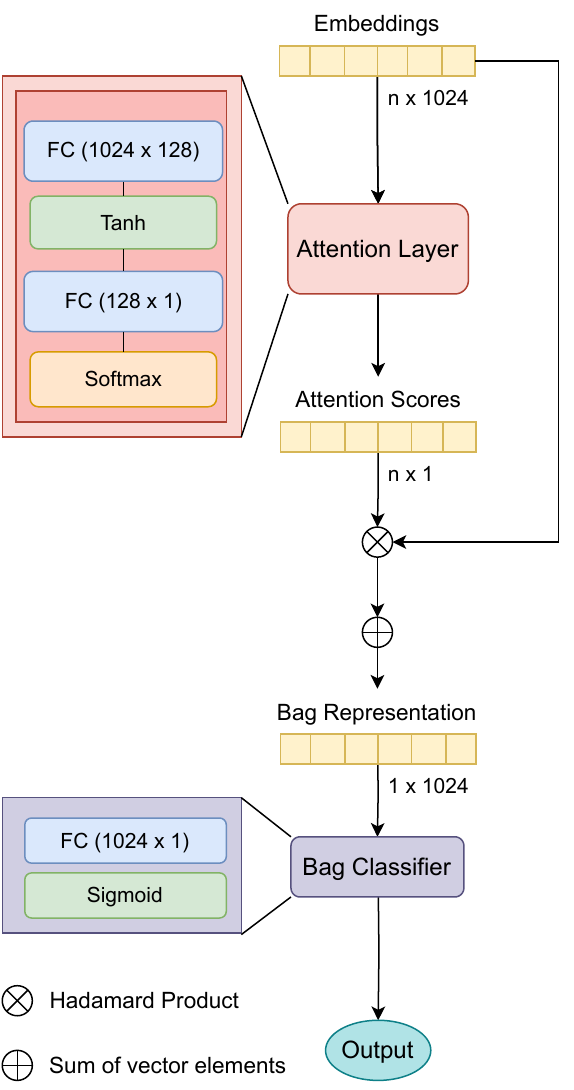}}
           \centering
           \hspace{4em} 
           \subfigure[Additive MIL Architecture.]{\label{fig:AdMIL}\includegraphics[width=0.26\textwidth]{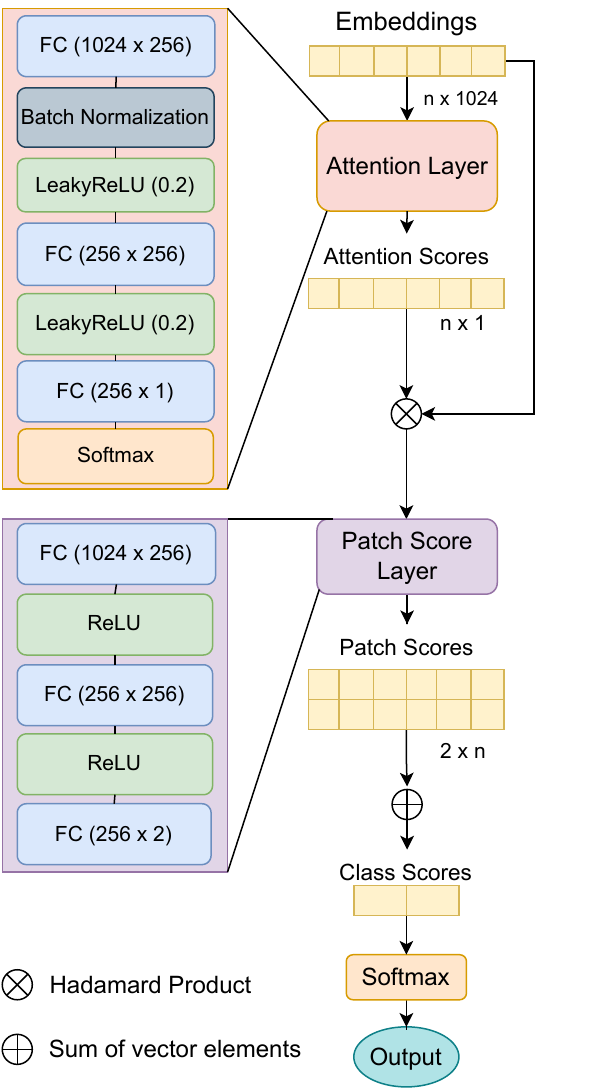}}   
            \caption[Models' Architectures.]{Models' Architectures. The attention layers are composed of fully-connected (FC) layers, followed by activation functions. 
            The AMIL (a) uses a tanh as its activation function, while the AdMIL (b) uses LeakyReLU. At the end of the attention layers, the results for each embedding are passed to a softmax to produce the final attention scores.}
            \label{fig:Architectures}
\end{figure}

The models' original architectures used in this work are represented in Figure~\ref{fig:Architectures}. The original AMIL model is composed of a feature extractor module, followed by the attention module which produces attention scores for each instance, and the final bag classifier module which takes
the sum of all instances multiplied by their respective scores and outputs the final bag classification.
Similarly to the AMIL architecture, the AdMIL model is composed of a feature extractor module
and an attention module. However, as mentioned before, instead of aggregating the patch embeddings through a weighted sum
and then passing the result through a final bag classifier, AdMIL passes each patch embedding multiplied by its 
attention score through a patch classifier, the Patch Score Layer. This classifier produces a patch score for each class. 
These scores are not bounded by any values and can be either
negative logits (inhibitory scores) or positive logits (excitatory scores). 
In the end, the scores are added together to produce 
a final bag score that is passed through a softmax, returning the final probability for each class.

By comparing the attention scores produced by the two models, we noticed that the use of the LeakyReLU 
activation function on the attention layer of the AdMIL 
architecture produced narrower attention scores, which might not be ideal, 
as it makes the model focus on a limited part of the input.  We decided to experiment on an additional model that uses the 
architecture of AdMIL, but with the attention layer of AMIL.

\section{Results}
\subsection{Training Methods}

We split each dataset into a training set (80\%) and test set (20\%). For each task and model, we performed 5-fold
cross-validation on the training set in order to prevent overfitting. The test set was then used for external validation,
with the model extracted from the fold that obtained the best results.

For all the models, the loss function used was the binary cross-entropy loss, defined as
\begin{equation}
    BCE = -(Y \times \log(P(Y)) + (1 - Y) \times \log(1 - P(Y))),
\end{equation}
where $Y$ is the positive class. For optimization of the loss, we used the Adam optimiser. 
For some of the models, we employed a cosine annealing learning rate scheduler.

\subsection{Datasets}
We chose two projects from TCGA (The Cancer Genome Atlas)~\cite{tcga} to be analyzed in this work: TCGA-BRCA (Breast Invasive Carcinoma) and TCGA-LUSC (Lung Squamous Cell Carcinoma).

For the tumor detection task, we only used flash-frozen slides.  Even though frozen specimens are less suitable for computational 
analysis when compared to formalin-fixed paraffin-embedded (FFPE)  slides, we decided to 
build our dataset with these because of the lack of FFPE slides in the TCGA containing only 
healthy tissue. For this task, we focused on 5x magnification tiles, since this is the magnification level that pathologists typically use when searching for tumors.

For gene mutation detection, we used FFPE Slides, since the unbalanced distribution was no longer a problem and these slides provided better 
results and training performance. Furthermore, the feature extractor we used for the tiles, Kimianet~\cite{riasatian2021fine}, was trained with 
FFPE slides, so conforming to the same would lead to better results. For this task, 
we built three datasets at three different magnification levels: 5x, 10x, and 20x, in order to better understand at which
magnification the models could best detect correlations between the mutation and the morphology of the tissue. 

Due to the large size of FFPE slides and to save storage space and time, 
we performed a random sampling of tiles for these slides, depending on the magnification level.
Moreover, while slide-level tumor presence labels were available for the first task, gene expression labels are only available at case-level (patient-level), presenting some challenges. We assumed 
that not only will the mutation be present in all diagnostic slides from a patient labeled as positive, but also that it 
would cover enough tissue to be captured in the tiles sampled in our dataset.

\subsubsection{TCGA-BRCA}
The TCGA-BRCA is composed of 1098 cases.
It contains 1133 FFPE Slides and 1978 flash-frozen slides across those cases. 
This dataset was used for the gene mutation task. We focused on the mutations of the TP53 gene since this gene shows a greater 
number of mutated cases from those tested for simple somatic mutations (331 of 969 cases), allowing us to build a balanced dataset. 
For this task, we have 349 positive slides and 670 negative slides. We chose an equal number of 
positive and negative WSIs, and after filtering inadequate slides through the processing pipeline, we ended up with a total of 662 slides, 331 labeled positive and 331 labeled negative. 

\subsubsection{TCGA-LUSC}
The TCGA-LUSC is a dataset for lung squamous cell carcinoma. The size of this dataset is fairly small when compared with 
TCGA-BRCA, with 504 cases, containing 512 diagnostic slides and 1100 tissue slides across those 
cases. 
For the tumor detection task, we have 753 positive slides and 347 negative slides.
Its class distribution for the tumor detection task is not too imbalanced for our purposes. We chose an equal number of positive and negative slides, ending up with a dataset composed of 694 slides. Its reduced number of slides, as well as the presence of more artifacts makes this type of cancer
more challenging to work with. The TCGA-BRCA slides always have a tumor percentage 
of 90\% at least, while in the case of TCGA-LUSC, the distribution of tumor percentage is more balanced. Therefore, we decided that 
for this task, a dataset built with slides from TCGA-LUSC is preferable for validating the ability of the model to generalize with different slides.

\subsection{WSI Preprocessing Pipeline}

WSIs are often too large to be directly processed by Deep Learning models.
Slides can occupy over 6 GB of memory, so having a dataset of raw slides on disk is not feasible in our case.
To store our dataset locally, we decided to focus on one magnification at a time. 
All our models were trained in separate magnification levels, which guaranteed that this would not be an issue.
For some magnifications, the total of patches might exceed the storage space available.
To overcome this issue, we developed a pipeline that fetches WSI tiles,  processes, filters, efficiently encodes them, and saves
the resulting embeddings and relevant metadata on HDF5 files. This pipeline is explained 
in the following sections.

\subsubsection{WSI Metadata Extraction}\label{WSI_Metadata_Extraction}
We start by extracting information about each slide: id, labels type of slide, and microns per pixel (mpp) at which the slide was originally scanned.
In the end, all the metadata extracted is saved to a CSV (comma-separated values) file for further processing.

\subsubsection{Patch Fetching and Pre-processing}\label{{DatasetPipeline}}
Whole slide images can have a lot of patches consisting exclusively of background or artifacts that make them unusable for training our models. Fetching all these unnecessary patches and checking them one by one becomes inefficient and time-consuming.
To avoid this we take advantage of WSI's multi-scale property to avoid  fetching patches that will definitely contain only background.

The tile at the thumbnail level is initially fetched. Otsu's thresholding~\cite{otsu1979threshold} is applied to it, 
as well as a close morphological operation to filter noise. We then extract the resulting black pixel coordinates,
obtaining a set of pixels $P$, the pixels that correspond to tissue.

In the acquisition of a WSI, there are often unintentional artifacts due to manual tissue
preparation, staining, and scanning hardware, as well as pathologists' annotations. 
To mitigate the number of tiles containing artifacts as well as remove some 
background pixels that might have passed through the previous filter, 
the color of each pixel $p \in P$ was compared to the average color of $P$, 
by calculating their Euclidean distance and comparing it with a pre-defined threshold. 
The coordinates of the pixels that fulfilled this condition were then stored and used to 
calculate the corresponding coordinates of the patches at the desired magnification, using 
the hierarchical properties of WSI and the metadata extracted from the step in~\ref{WSI_Metadata_Extraction}.

Because these filters were applied at the thumbnail level, some of the tiles were still not suitable for the final dataset.
After fetching each tile, we checked if its size in image coordinates was 512 $\times$ 512 px. If it was smaller, it was padded accordingly 
with the average background color. The percentage of tissue present in the tile was then 
calculated and compared with a threshold. If it does not contain enough tissue, the image is discarded.

\begin{figure}[hbtp]
    \centering
        \includegraphics[width=0.8\textwidth]{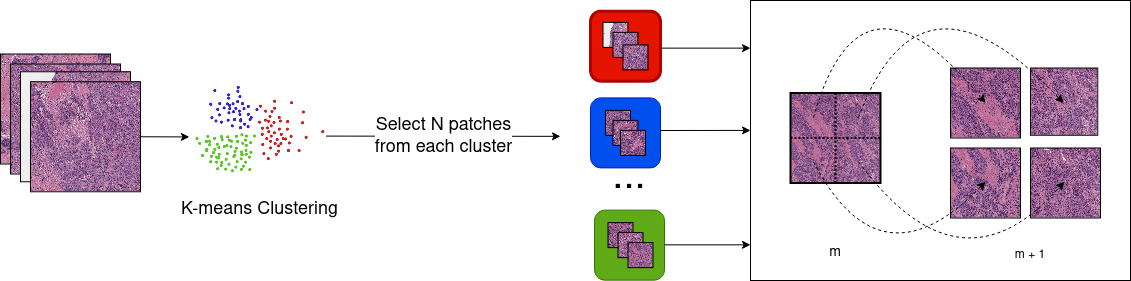}
    \caption[Sampling method for 10x and 20x magnification.]{Sampling method for 10x and 20x magnifications. K-means clustering is applied to the set of tiles chosen from the previous magnification $m$.
    N tiles from each cluster are selected and the corresponding tiles at the magnification desired ($m$ + 1) are then fetched.}
    \label{fig:clustering}
\end{figure}

For Gene Mutation Detection, due to the size of the FFPE slides and the multiple magnification levels used,
we applied random sampling to the tiles:
\begin{itemize}
    \item At 5x magnification, we sampled 60\% of the filtered tiles, when 
    their number was greater than a certain limit; otherwise, we used all the tiles.
    \item In the case of 10x and 20x magnification,
    we applied clustering on the tiles from the previous magnification and sampled a chosen number n of tiles 
    per cluster.
    We then proceeded to use the hierarchical properties of WSI to extract the corresponding 
    tiles at the desired magnifications. For instance, we performed K-means clustering on the tiles at 5x magnification, 
    sampled a maximum of 20 tiles from each cluster, and proceeded to extract, for each tile, the corresponding 
    tiles at 10x magnification (Figure~\ref{fig:clustering}).  
\end{itemize}

\subsubsection{Feature Extraction}
As mentioned previously, embeddings were created from the tiles. For this, we chose KimiaNet\cite{riasatian2021fine}, a Densenet 121 pretrained for WSI tumor subtype classification,
that produces embedding vectors with a length of 1024. This model was trained exclusively on FFPE slides.

We also performed two data augmentations per tile, composed of random HED stain perturbation, 
Gaussian noise addition, rotations, and horizontal and vertical flips. Embeddings were generated for these augmented tiles as well. The embeddings are then saved to an HDF5 file, along with the corresponding relevant metadata, such as the 
coordinates of the patches and labels. This file can then be quickly read to memory during the training process of 
the models.

Due to the size of the dataset, immediately converting each 512 $\times$ 512 pixels patch to an
embedding of length 1024 saves storage space (with a decrease in size close to a thousand-fold) and time 
for training the model and allows us to have the whole dataset locally, instead of having to 
fetch the data each time we needed to train or fine-tune the models. Furthermore, by having the 
slides represented in feature space immediately, as opposed to pixel space, we were able to fit 
all patches in a slide into GPU memory concurrently, which is especially useful for multiple instance learning 
approaches. 

\subsection{Classification and RoI Detection Results}
In this section, we present the results obtained for our classification tasks, as well as some examples of the RoIs produced by the models.
After fine-tuning, we performed, for each model, task, and magnification level, five independent runs where the datasets
were randomly split with different seeds. Tables~\ref{tab:tp53_5x_metrics} and \ref{tab:gene_task} display the average and standard deviation
of the Areas under the Curves (AUCs) obtained on the test set for those five runs. 
We also present graphs displaying the ROC curves of one of these runs for each model, task and magnification.

Table~\ref{tab:tp53_5x_metrics} shows the AUC results for the tumor detection task at magnification 5x. Figure~\ref{fig:LUSC_Heatmaps} shows an example of the heatmaps produced for the tumor detection task in a slide at the same magnification level.
Table~\ref{tab:gene_task} presents the gene mutation detection task AUC results at magnification levels 5x, 10x and 20x. Figures~\ref{fig:10x_Heatmaps} and \ref{fig:20x_Heatmaps} present examples of heatmaps produced for the gene mutation detection task from patches of magnification 10x and 20x. Due to the poor classification results obtained at magnification 5x, we did not include heatmaps from this level, since we concluded that they would not be meaningful.

For the AMIL model we only have the attention scores, corresponding to the patches that were considered the most relevant for the final prediction.
Regarding the original AdMIL and our version, we have the patch attention scores 
produced by the attention layer, as well as the excitatory and inhibitory patch scores, that indicate positive and negative contributions for the final prediction, respectively. These final scores were passed through a sigmoid to scale the logits to values between 0 and 1, where values in the interval ]0, 0.5[ indicates a negative contribution and values in [0.5, 1[ a positive contribution.
In the case of attention heatmaps, we show a continuous colormap. For the inhibitory/excitatory scores, we only use two colors, one for the excitatory patches (Red) and other for the inhibitory patches (Blue).

\begin{table}[H]
\centering
    \begin{tabular}{|l|c|}
        \hline
        \textbf{Model Architecture} & \textbf{AUROC} \\
        \hline
        Attention MIL (AMIL) & $\mathbf{0.971 \pm 0.015}$ \\
        Additive MIL (AdMIL)& $\mathrm{0.955 \pm 0.012}$ \\
        Our AdditiveMIL & $\mathrm{0.963 \pm 0.020}$\\
        \hline
    \end{tabular}
\captionsetup{justification=justified}
\caption{Models' Performance on the tumor detection task. After fine-tuning, we performed the external validation on the test set five times for each model and calculated the average AUC of those 5 runs. The original architecture of AMIL obtained the best performance, followed by our version of AdMIL.}\label{tab:tp53_5x_metrics}
\end{table}

\begin{table}[H]
    \centering
    \begin{small}
        \begin{tabular}{cl*{3}{S}}
            \toprule
            Magnification Level &  Model Architecture  & \text{AUROC}  \\
            \midrule
            \multirow{3}{*}{5x}  
                & Attention MIL (AMIL) & $\mathbf{0.605 \pm 0.020}$   \\
                & Additive MIL (AdMIL) & $\mathrm{0.575 \pm 0.038}$   \\
                & Our Additive MIL   & $\mathrm{0.512 \pm 0.034}$     \\
            \cmidrule(lr){1-3}
            \multirow{3}{*}{10x}  
                & Attention MIL (AMIL) & $\mathbf{0.711 \pm 0.033}$ \\
                & Additive MIL (AdMIL) & $\mathrm{0.624 \pm 0.046}$ \\
                & Our Additive MIL   & $\mathrm{0.547 \pm 0.045}$ \\
            \cmidrule(lr){1-3}
            \multirow{3}{*}{20x}  
                & Attention MIL (AMIL) & $\mathbf{0.704 \pm 0.040}$ \\
                & Additive MIL (AdMIL) & $\mathrm{0.624 \pm 0.033}$ \\
                & Our Additive MIL   & $\mathrm{0.642 \pm 0.029 }$ \\
            \bottomrule
        \end{tabular}
    \end{small}
    \captionsetup{justification=justified}
    \caption{Models' Performance for the gene mutation detection task for the three magnification levels explored (5x, 10x and 20x). After fine-tuning, we performed the external validation on the test set five times for each model and calculated the average AUC of those five runs. The AMIL architecture always obtained the best performance. In general, the models obtained a better performance for higher magnification levels, with our version of AdMIL obtaining a better performance than the original at magnification level 20x.}\label{tab:gene_task}
  \end{table}

\begin{figure}[htbp]
\centering
        \subfigure[AMIL model]{\label{fig:ROC_TUMOR_Attention}\includegraphics[width=0.28\textwidth]{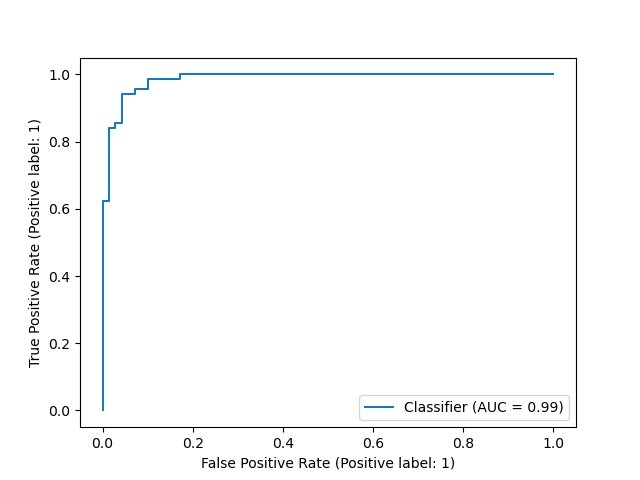}}
        \subfigure[AdMIL model]{\label{fig:ROC_TUMOR_AdMIL}\includegraphics[width=0.28\textwidth]{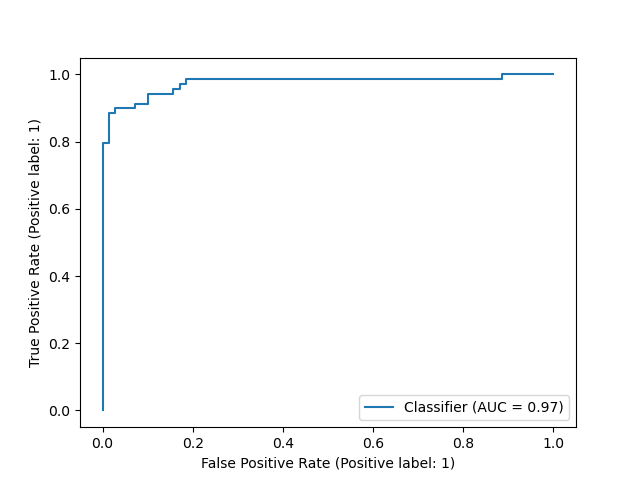}}
        \subfigure[Our AdMIL model]{\label{fig:ROC_TUMOR_Mod}\includegraphics[width=0.28\textwidth]{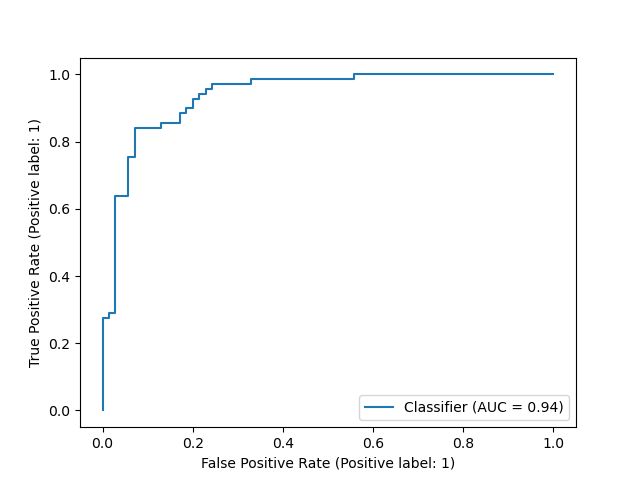}}
        \caption{ROC curves for one of the runs of the Tumor Detection Task (5x magnification)}
        \label{fig:ROC_CURVES_TUMOR}
\end{figure}

\begin{figure}[htbp]
\centering
        \subfigure[AMIL model]{\label{fig:ROC_TP53_5x_Attention}\includegraphics[width=0.28\textwidth]{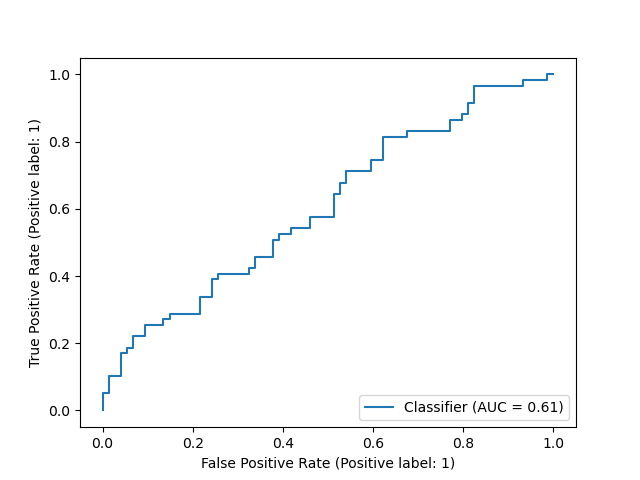}}
        \subfigure[AdMIL model]{\label{fig:ROC_TP53_5x_AdMIL}\includegraphics[width=0.28\textwidth]{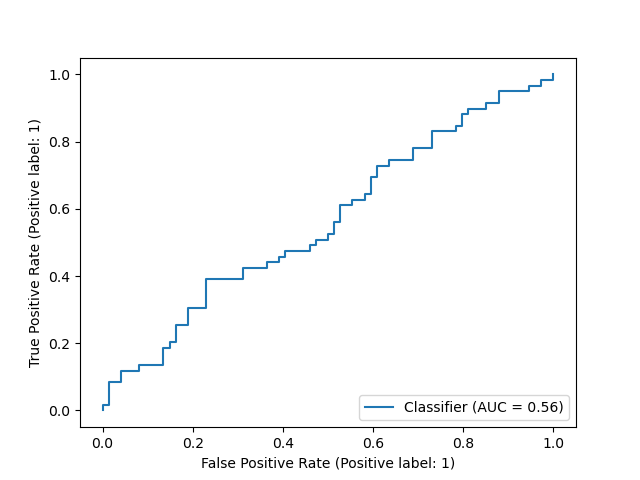}}
        \subfigure[Our AdMIL model]{\label{fig:ROC_TP53_5x_Mod}\includegraphics[width=0.28\textwidth]{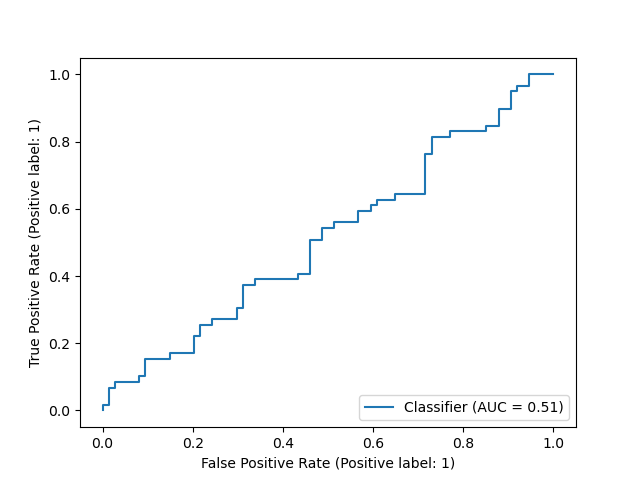}}
        \caption{ROC curves for one of the runs of the TP53 mutation Detection Task (5x magnification)}
        \label{fig:ROC_CURVES_TP53_5x}
\end{figure}

\begin{figure}[htbp]
\centering
        \subfigure[AMIL model]{\label{fig:ROC_TP53_10x_Attention}\includegraphics[width=0.28\textwidth]{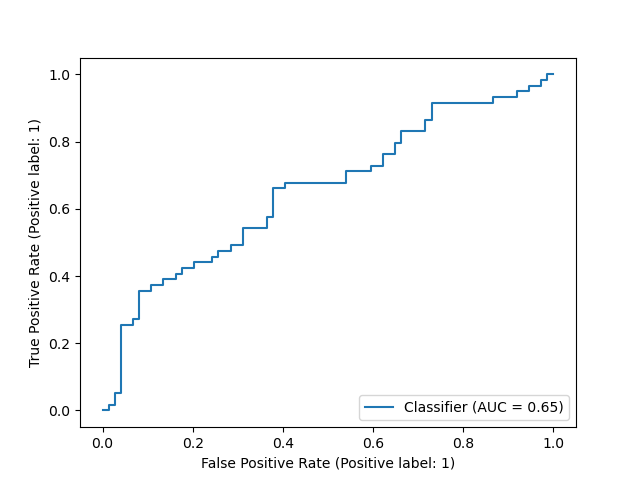}}
        \subfigure[AdMIL model]{\label{fig:ROC_TP53_10x_AdMIL}\includegraphics[width=0.28\textwidth]{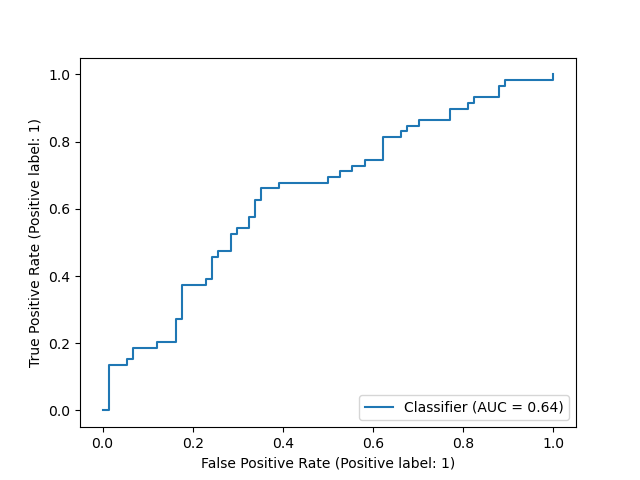}}
        \subfigure[Our AdMIL model]{\label{fig:ROC_TP53_10x_Mod}\includegraphics[width=0.28\textwidth]{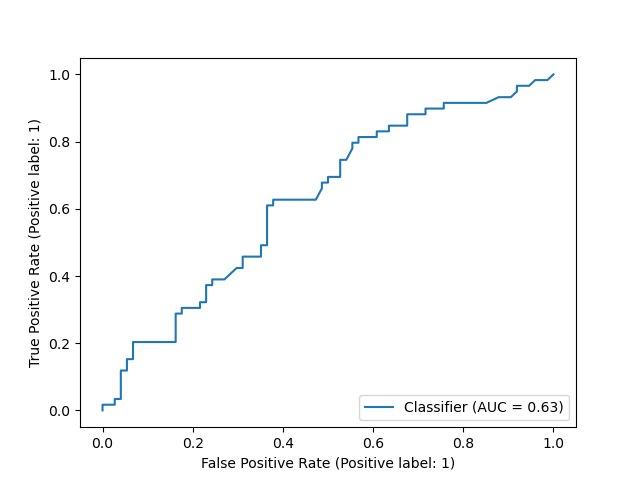}}
        \caption{ROC curves for one of the runs of the TP53 mutation Detection Task (10x magnification)}
        \label{fig:ROC_CURVES_TP53_10x}
\end{figure}

\begin{figure}[htbp]
\centering
        \subfigure[AMIL model]{\label{fig:ROC_TP53_20x_Attention}\includegraphics[width=0.28\textwidth]{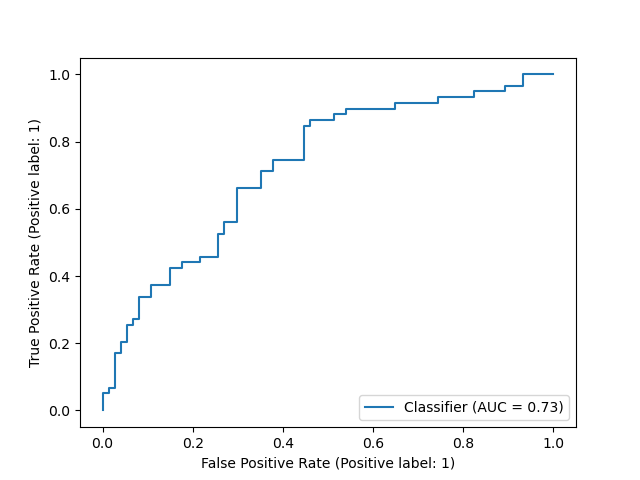}}
        \subfigure[AdMIL model]{\label{fig:ROC_TP53_20x_AdMIL}\includegraphics[width=0.28\textwidth]{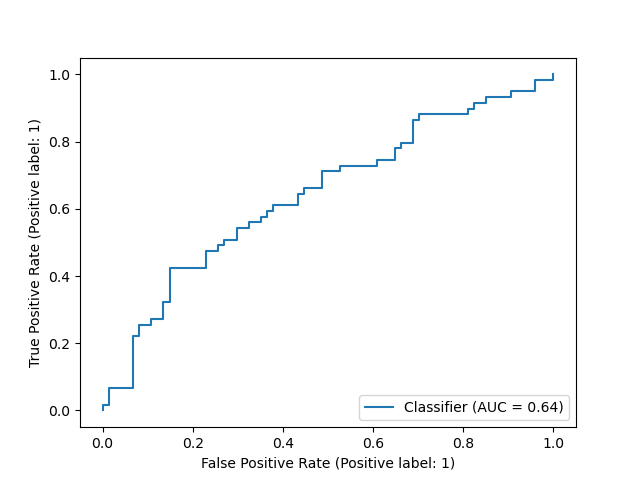}}
        \subfigure[Our AdMIL model]{\label{fig:ROC_TP53_20x_Mod}\includegraphics[width=0.28\textwidth]{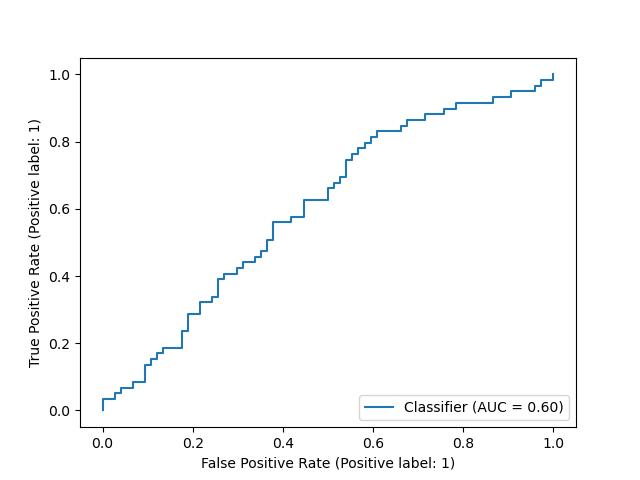}}
        \caption{ROC curves for one of the runs of the TP53 mutation Detection Task (20x magnification)}
        \label{fig:ROC_CURVES_TP53_20x}
\end{figure}




\begin{figure}[htbp]
\centering
        \subfigure[Original Slide]{\label{fig:Original_slide_LUSC_5x_1}\includegraphics[width=0.140\textwidth]{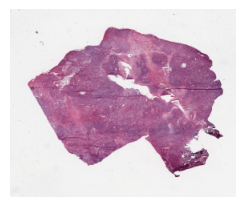}}
        \subfigure[AMIL Heatmap]{\label{fig:AMIL_heatmap_LUSC_5x_1}\includegraphics[width=0.140\textwidth]{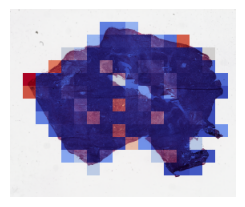}}
        \subfigure[AdMIL Scores Heatmap]{\label{fig:AdMIL_heatmap_LUSC_5x_1}\includegraphics[width=0.140\textwidth]{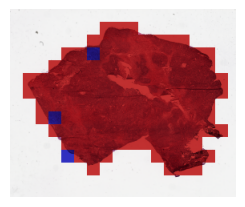}}
        \subfigure[AdMIL Attention Heatmap]{\label{fig:AdMIL_attention_LUSC_5x_1}\includegraphics[width=0.140\textwidth]{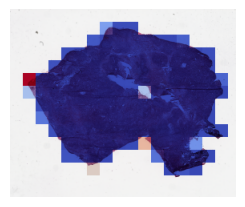}}
        \subfigure[Our AdMIL Scores Heatmap ]{\label{fig:ModAdMIL_heatmap_LUSC_5x_1}\includegraphics[width=0.140\textwidth]{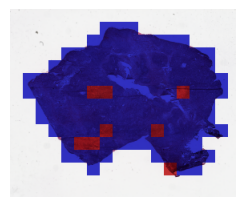}}
        \subfigure[Our AdMIL Attention Heatmap]{\label{fig:ModAdMIL_attention_LUSC_5x_1}\includegraphics[width=0.140\textwidth]{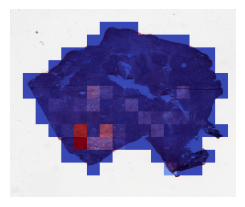}}
        \caption{Heatmaps produced for slide 05194610-c32d-44db-be55-c38652d547b8 for the tumor detection task at 5x magnification. (a) Slide at thumbnail level;
        (b) AMIL attention scores; (c) AdMIL's final patch scores; (d) AdMIL's attention scores;
        (e) Our version of AdMIL's final patch scores; (f) Our version of AdMIL's attention scores.
        }
        \label{fig:LUSC_Heatmaps}
\end{figure}

\begin{figure}[htbp]
    \centering
        \subfigure[Original Slide]{\label{fig:Original_slide_BRCA_10x}\includegraphics[width=0.14\textwidth]{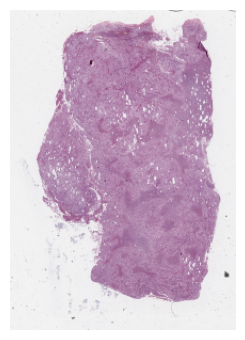}}
        \subfigure[AMIL Heatmap]{\label{fig:AMIL_heatmap_BRCA_10x}\includegraphics[width=0.14\textwidth]{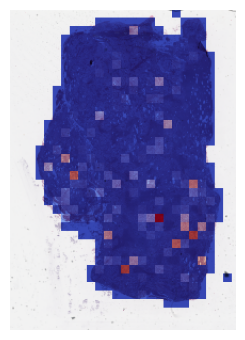}}
        \subfigure[AdMIL Heatmap]{\label{fig:AdMIL_heatmap_BRCA_10x}\includegraphics[width=0.14\textwidth]{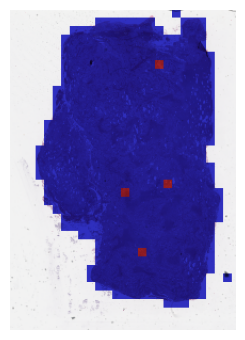}}
        \subfigure[AdMIL Attention]{\label{fig:AdMIL_attention_BRCA_10x}\includegraphics[width=0.14\textwidth]{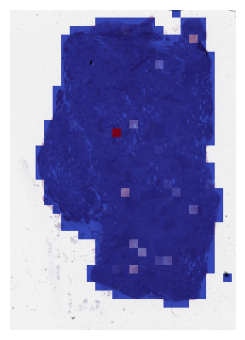}}
        \subfigure[Our AdMIL Heatmap]{\label{fig:ModAdMIL_heatmap_BRCA_10x}\includegraphics[width=0.14\textwidth]{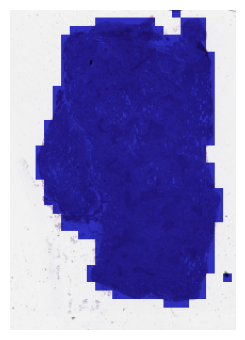}}
        \subfigure[Our AdMIL Attention]{\label{fig:ModAdMIL_attention_BRCA_10x}\includegraphics[width=0.14\textwidth]{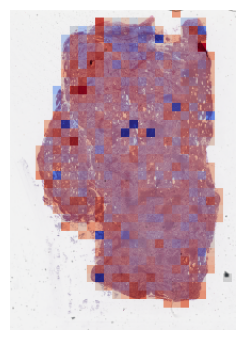}}
        \caption{
        Heatmaps produced for slide d42eb4ce-d300-46be-8b46-059a98c2ddb8 for the gene mutation detection task at 10x magnification. (a) Slide at thumbnail level;
        (b) AMIL attention scores; (c) AdMIL's final patch scores; (d) AdMIL's attention scores;
        (e) Our version of AdMIL's final patch scores; (f) Our version of AdMIL's attention scores.
        }
        \label{fig:10x_Heatmaps}
\end{figure}

\begin{figure}[htbp]
\centering
        \subfigure[Original Slide]{\label{fig:Original_slide_BRCA_20x}\includegraphics[width=0.14\textwidth]{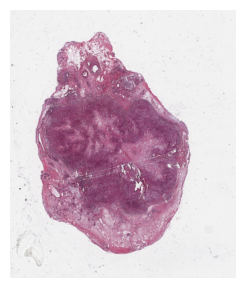}}
        \subfigure[AMIL Heatmap]{\label{fig:AMIL_heatmap_BRCA_20x}\includegraphics[width=0.14\textwidth]{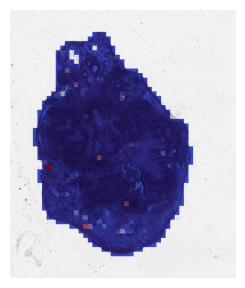}}
        \subfigure[AdMIL Heatmap]{\label{fig:AdMIL_heatmap_BRCA_20x}\includegraphics[width=0.14\textwidth]{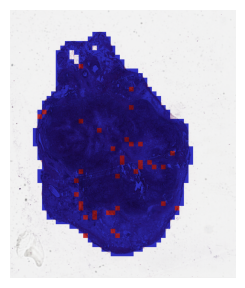}}
        \subfigure[AdMIL Attention]{\label{fig:AdMIL_attention_BRCA_20x}\includegraphics[width=0.14\textwidth]{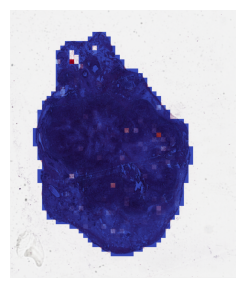}}
        \subfigure[Our AdMIL Heatmap]{\label{fig:ModAdMIL_heatmap_BRCA_20x}\includegraphics[width=0.14\textwidth]{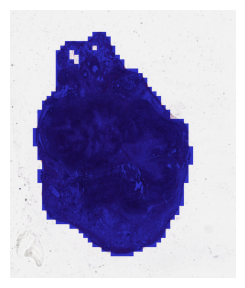}}
        \subfigure[Our AdMIL Attention]{\label{fig:ModAdMIL_attention_BRCA_20x}\includegraphics[width=0.14\textwidth]{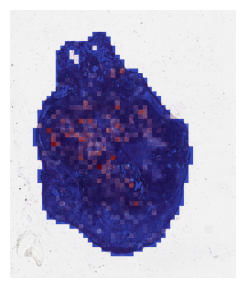}}
        \caption{Heatmaps produced by the models for slide cf1ee3be-f161-4614-b2dd-030a9adfc5fc for the gene mutation detection task at 20x magnification. (a) Slide at thumbnail level; (b) AMIL attention scores; (c) AdMIL's final patch scores; (d) AdMIL's attention scores;
        (e) Our version of AdMIL's final patch scores; (f) Our version of AdMIL's attention scores.
        }
        \label{fig:20x_Heatmaps}
\end{figure}

\section{Discussion}

\subsection{Tumor Detection Task}
The models performed on par with the state-of-the-art for this task, in terms of their ability to correctly classify WSI, with AUC values above 0.9 
(Table \ref{tab:tp53_5x_metrics}). 
The original article~\cite{Ilse2018} used two H\&E image datasets: the breast cancer dataset
and the colon cancer dataset. Comparing our results with the ones reported in this work, we obtained significantly better results than the first and similar results to the second. We should note however that these datasets are not only from a different cancer type but also composed of patches as bags, instead of slides.
Other works used the AMIL architecture and its own variations with datasets composed of tiles at higher magnifications~\cite{Li2021, shao2021transmil}. In general, we obtained a better AUC than previous reported work, which supports the fact that 5x magnification might be adequate for identifying tumors.

Considering that the dataset we used only had flash-frozen slides, that it presented some artifacts, and that it included varied percentages of tumors,
we can assume that the models learned to differentiate unrelated factors from the tumors present in the images. Moreover, these results
were obtained from tiles at 5x magnification, a level that displays tissue but not cells. This supports the hypothesis that, just like pathologists,
models can learn to identify tumor slides at the tissue level. 
It also reveals that the task in itself might not be challenging enough,  and exploring higher magnifications will not meaningfully improve the slides' classification ability. It was due to these results and the fact that this is the most common level used for tumor detection by pathologists that
we did not explore further magnification levels for this task.

By analyzing the attention scores produced, we observed that both the 
AMIL Architecture (Figure\ref{fig:AMIL_heatmap_LUSC_5x_1}) and our version of the AdMIL model
(Figure \ref{fig:ModAdMIL_attention_LUSC_5x_1}) 
produce sparser attention scores when compared with the original AdMIL (Figure\ref{fig:AdMIL_attention_LUSC_5x_1}) as we previously
predicted. This behavior might be especially important 
if the tumor regions (or the desired RoIs) appear more scattered. On the other hand,
for morphologies that are usually more focused on a specific patch, the attention layer from the original AdMIL
could be preferable.

On the other hand, when analyzing the excitatory/inhibitory scores produced by the Additive MIL framework, 
the scores produced by the original AdMIL model 
tend to be positive for most of the patches (Figure\ref{fig:AdMIL_heatmap_LUSC_5x_1}), while
the ones produced by our version of AdMIL only highlight a small 
portion~(Figure\ref{fig:ModAdMIL_heatmap_LUSC_5x_1}). 
This behavior was observed for most of the slides.
Since the only difference in the models' architecture is their attention layers, we conclude that sparser attention
scores tend to make the instance classifier consider only a small subset of patches as excitatory. 
Depending on the results acquired from a future evaluation of the heatmaps' relevance,
this might support the hypothesis that attention scores do not necessarily highlight the desired RoIs and might not be enough for 
meaningful WSI painting.

It is also interesting to notice that the AMIL attention scores tend to focus more on the edges of tissue (Figure~\ref{fig:AMIL_heatmap_LUSC_5x_1}). This might mean that  the tumor is present in the borders of the tissue, but it might also reveal some bias in the model for patches that include some 
percentage of background. Due to the magnification used, and the fractures that appear in flash-frozen slides, there was a significant amount of patches that contained background, when compared to higher magnifications and it is possible that this conditioned the model's learning ability.

Further evaluation of the RoI detection has to be done, especially a qualitative evaluation of the heatmaps obtained. We would need the help of pathologists
to confirm that the patch scores that these models obtain are indicators of tumor regions. Furthermore, while the AMIL model showed a slightly better AUC than the other models, this does not necessarily mean that its heatmaps are more relevant or helpful for pathologists.
The patch scores obtained from the other two models, due to their inhibitory/excitatory nature, might provide greater insight into tumor presence in the slides.
Due to time constraints and the non-existence of ground truth heatmaps, this evaluation is left as future work.

\subsection{Gene Mutation Detection Task}

For the magnification level 5x, the models' performance was poor (Table\ref{tab:gene_task}). The AMIL model showed the best performance, with an average AUC of 0.605. The hypothesis that was already supported by the results 
in previous work~\cite{wang2021prediction} is still supported here: tiles at 5x magnification, a level at which only tissue is visible, cannot show discernible evidence of TP53 mutations for most cases. 
Therefore, regarding the model's ability to identify mutations for this gene, it is not possible to draw conclusions from this magnification level.

At level 10x, we obtained far better AUC results (Table\ref{tab:gene_task}). The AMIL model still shows the best performance in terms of its average AUC (0.711), but the AdMIL model also presents a reasonable score (0.624). Our version of AdMIL, however, did not improve considerably for most runs, with an average score of 0.547.

For this task, at this level, the number of patches highlighted in heatmaps is far scarcer. In the case of the AMIL model,
unlike the previous task, the attention mechanism focused more on the inner patches (Figure~\ref{fig:AMIL_heatmap_BRCA_10x}). This could be due to the fact that it is looking for a different pattern in the morphology that might not be present in the borders, but it is also true that, at these magnification levels, the number of patches with background is more scarce, unlike the previous task.

The original AdMIL model focused on a very short number of patches (Figure~\ref{fig:AdMIL_heatmap_BRCA_10x}). Even though it classifies slides as showing signs of TP53 mutation, at this magnification level, it does not seem to be able to point to relevant regions, 
only highlighting individual patches that are distant and isolated from each other. 
Just like in the tumor detection task, its attention mechanism only focused on a small number of patches (Figure~\ref{fig:AdMIL_attention_BRCA_10x}).

Our version of AdMIL, on the other hand, is unable to identify any excitatory patches from the majority of WSI (Figure~\ref{fig:ModAdMIL_heatmap_BRCA_10x}). 
This might have to do with the fact
that its attention mechanism is not able to make a notable distinction between patches, attributing very similar attention scores to all of them (Figure~\ref{fig:ModAdMIL_attention_BRCA_10x}). 
This also explains the low average AUC obtained by this model at this magnification level.

At level 20x, the results did not change by much for AMIL and AdMIL, but improved a lot for our modified version of AdMIL, with an AUC score of 0.642. 
For this level, the heatmaps produced also did not show much change from the ones at magnification 10x. However, it is interesting to notice that the attention scores produced by our version
of AdMIL, unlike the previous level, focused around a specific region on the WSI (Figure~\ref{fig:ModAdMIL_attention_BRCA_20x}). However, this model continues to produce only inhibitory scores.

The AUC scores obtained by our models for levels 10x and 20x are comparable with 
the ones obtained by previous work~\cite{guo2023robust, reisenbuchler2022local}, although we have used simpler architectures. 
However, this comparison should be taken carefully, since the models were trained on different datasets (with different methods for tile preprocessing)
and often for a different type of gene mutation.

In general, the quantitative results for this task are not as good as those from the tumor detection task.
The heatmaps produced do not seem to highlight a significant number of patches. This might be because signs 
of TP53 mutation might appear more isolated in these slides. Just like for the previous task, further evaluation of the heatmaps
by specialists would have to be done to reach a definite conclusion.

We speculate that the models' ability to learn patterns related to the TP53 mutation might have been affected by the dataset used.
Although we only used FFPE slides, which are typically better for WSI computational analysis, 
the fact that, due to time constraints, our WSI bags were built using random sampling might have reduced the quality of their representation of the slide.
Furthermore, we assumed that all slides belonging to a positive patient are positive as well. All these factors might have introduced too much noise to the dataset.

Beside the possibility of too much noise, we come to the conclusion that detecting patterns of TP53 mutation
on digital slides with MIL and attention mechanisms is a much more challenging task, and the results are not as good 
as when detecting the presence of tumors.

\section{Conclusions}

In this work, we study the performance of multi-instance learning (MIL) frameworks with attention mechanisms for WSI classification and virtual staining of breast tumors. We explored two distinct approaches that identify different morphology proxies for patch (tile) classifiers. These two frameworks were used in a weakly-supervised application to tumor detection and TP53 mutation detection in Breast Carcinoma Lung and Squamous Cell Carcinoma. We found that it was far easier to identify Regions of Interest (RoIs) recognizing tumor vs non-tumor even at low resolution (AUC > 0.95), than it was to classify TP53 mutated vs non-mutated (AUC < 0.71). The observation that higher resolutions (20x) worked better to identify RoI for mutation was by itself not a surprise, but the opportunities to hypothesize novel morphological interpretations emerged as the main result from the work reported here. Similarly, when we explored new modifications of the established multi-instance learning method by altering its original attention layer, the most interesting result was not improved accuracy but the attention the model placed on different morphological features ( i.e. "slide painting"). Specifically, the results described in Figures 3, 4, and 5 illustrate the opportunities for interactive exploration of recurring morphologies for their role in cancer etiology. 

\section{Acknowledgements}
This work was funded in part by the National Cancer Institute (NCI) Intramural Research Program CAS\#10901 (EpiSphere), the Recovery and
Resilience Fund towards the Center for Responsible AI (Ref. C628696807-00454142), and the financing of the Foundation for Science and
Technology (FCT) for INESCID (Ref. UIDB/50021/2020).

The open source code used in this project is publicly available at \href{https://github.com/timafonso/WSI\_MIL\_ROI}{https://github.com/timafonso/WSI\_MIL\_ROI}

\section{Author Declaration}

There is no conflict of interest to report.

\bibliographystyle{cas-model2-names}

\bibliography{cas-refs}





\end{document}